\documentclass[11pt,a4paper]{article}
\usepackage{emnlp2016}
\usepackage{times}
\usepackage{url}
\usepackage{latexsym}
\usepackage{amssymb}
\usepackage{amsmath}
\usepackage{lingmacros}
\usepackage{multicol}
\usepackage{color}
\usepackage{graphics}
\usepackage{graphicx}

\usepackage{subfigure}
\usepackage{algorithm} 

\pdfoutput=1

\usepackage{xcolor}

\usepackage{algpseudocode}
\newcommand{\hgfc}{\textsc{hgfc}}    

\definecolor{shadecolor}{rgb}{0.9,0.9,0.9}
\definecolor{olive}{rgb}{0.3, 0.4, .1}
\definecolor{dgreen}{rgb}{0.,0.6,0.}
\definecolor{gold}{rgb}{1.,0.84,0.}
\definecolor{JungleGreen}{cmyk}{0.99,0,0.52,0}
\definecolor{BlueGreen}{cmyk}{0.85,0,0.33,0}
\definecolor{RawSienna}{cmyk}{0,0.72,1,0.45}
\definecolor{Magenta}{cmyk}{0,1,0,0}

\newcommand{\citet}[1] {\newcite{#1}}
\newcommand{\citep}[1] {\cite{#1}}
\newcommand{\shorten}[1]{}

\emnlpfinalcopy

\title{Psychologically Motivated Text Mining}

\author{Ekaterina Shutova\\
	    Computer Laboratory\\
	    University of Cambridge\\
	    {\tt es407@cam.ac.uk}
	  \And
	Patricia Lichtenstein\\
	    Dept. of Cognitive and Information Sciences\\
	    University of California, Merced\\
	    {\tt tricia1@uchicago.edu }}

\date{}

\begin{document}
\maketitle
\begin{abstract}
 Natural language processing techniques are increasingly applied to identify social trends and predict behavior based on large text collections. Existing methods typically rely on surface lexical and syntactic information. Yet, research in psychology shows that patterns of human conceptualisation, such as metaphorical framing, are reliable predictors of human expectations and decisions. In this paper, we present a method to learn patterns of metaphorical framing from large text collections, using statistical techniques. We apply the method to data in three different languages and evaluate the identified patterns, demonstrating their psychological validity.
\end{abstract}

\section{Introduction}

With the rise of blogging and social media, applying text mining techniques to aid political and social science has become an active area of research in natural language processing (NLP) \cite{grimmer2013}. NLP techniques have been successfully used for tasks such as estimating the influence of politicians \cite{fader2007}, predicting voting patterns \cite{gerrish2011} and political affiliation \cite{pennacchiotti2011}. Such methods typically rely on surface lexical and syntactic cues, rather than analysing patterns of conceptualisation and framing of social and political issues. Framing is, however, widely studied in political science, linguistics and cognitive psychology \cite{Lakoff1991,Tannen,Entman2003} as a way of reasoning about an issue by selecting and emphasizing its facets that reinforce a particular point of view. 
  Metaphor is a particularly apt framing device, as it exposes the desired aspects of the issue, while seamlessly concealing the less desired ones \cite{Lakoff1991,BeataACL2010,LakoffWehling}. For instance, discussing \textit{war} as a \textit{competitive game} emphasizes the victory vs. defeat aspect of war, while neglecting its human cost. Sports metaphors have thus been often used by politicians seeking to arouse a pro-war sentiment in the public \cite{Lakoff1991}. 

Psychologists \newcite{ThibodeauBoroditsky} investigated how metaphors affect our decision-making. 
In their experiment, two groups of subjects were primed by two different metaphors for \textit{crime}: \textit{crime is a virus} vs. \textit{crime is a beast} and then asked how crime should be tackled.
They found that the first group tended to opt for preventive measures and the second group for punishment-oriented ones. According to the authors, their results demonstrate the influence that metaphors have on how we conceptualize and act with respect to societal issues. 
 This in turn suggests that the metaphors we use can serve as a predictor of our social, political and economic decisions. Therefore, a text mining system aiming to gain an understanding of social trends across populations or their change over time, needs to identify subtle but systematic linguistic differences, expressed both literally and metaphorically. 

 In this paper, we propose a method for 
 large-scale identification of metaphorical framing patterns in text corpora. Metaphorical expressions arise in the presence of systematic metaphorical associations, or \textit{conceptual metaphors}, mapping one concept or domain to another \cite{LakoffAndJohnson}. For instance, when we talk about ``\textit{curing} juvenile delinquency'' or ``\textit{diagnosing} corruption'', 
   we view \textit{crime} (the \textit{target} concept) as a \textit{virus} or a \textit{disease} (the \textit{source} concept). 
 Our method uses clustering techniques to generalise such metaphorical associations based on the metaphorical use of language in a large text corpus. Specifically, we use a hierarchical soft clustering method -- hierarchical graph factorization clustering ({\hgfc}) \cite{yu2006soft} -- to learn a graph of concepts from the data and to identify patterns of inter-conceptual association in this graph. 
To obtain the graph, we cluster frequent nouns in the corpus using the verbs they co-occur with as features. Our expectation is that the verbs that systematically occur with both the source domain nouns (e.g. ``cure disease'') and the target domain nouns (e.g. ``\textit{cure} crime'') would allow the system to establish a connection between the two domains, providing evidence of metaphorical framing.

The method is fully unsupervised and relies on uncovering the patterns of systematic use of metaphor in linguistic data. It can thus be applied to any text corpus, domain or language, without any need for manual annotation. We apply the method to large corpora in three languages -- English, Russian and Spanish. The method identifies interesting differences in metaphorical framing in these corpora. We validated these differences in a behavioural experiment and established that the system accurately predicts behavioural data on human judgements of economic change. 
 Besides providing a new set of techniques for text mining applications, this method can also inform and scale-up research in experimental psychology, based on data-driven evidence rather than introspection.

\section{Related work}

NLP techniques have been successfully used for a number of tasks in political science, including automatically estimating the influence of particular politicians in the US senate \cite{fader2007}, identifying lexical features that differentiate political rhetoric of opposing parties \cite{monroe2008}, predicting voting patterns of politicians based on their use of language \cite{gerrish2011}, and predicting political affiliation of Twitter users \cite{pennacchiotti2011}. Other approaches \cite{paul2009,ahmed2010,fang,qiu2013,gottipati2013} detected the contrasting perspectives on a set of topics attested in distinct corpora using LDA topic modelling. 
Some works focused on subjectivity detection, identifying opinion, evaluation, and speculation in text \cite{wiebe2004}\shorten{.
There is a large body of work on identifying such opinions} and attributing it to specific people \cite{awadallah2011,abujbara2012}.
While successful in their tasks, these methods rely on surface linguistic cues, rather than generalising patterns of human association and conceptualisation, which limits the information they discover to that explicitly stated. 

In the meantime, much work on framing in political science and linguistics has shown that systematic variations in the use of metaphor across communities is a rich source of information about the differences in world views \cite{LakoffWehling,Shaikh2014,DiazVera,kovecses,DeignanPotter2004,Stefanowitsch2004,Musolff}. 
 For instance, \newcite{LakoffMoralPolitics} discuss how two conflicting instantiations of the \textsc{nation is a family} metaphor, using \textit{nurturing-parent} vs. \textit{strict-father} family models, explain the liberal-conservative divide in the US politics. 
 \newcite{LakoffWehling} show that the two models are consistent with both the parties' rhetoric and their policies, and are a reliable predictor of liberal vs. conservative values.  Some works \cite{Charteris-Black-Ennis,barcelona,matsuki,taylorandmbense}
  studied metaphor cross-linguistically and invariably found distinct patterns of metaphorical use across languages. 

The majority of computational approaches to metaphor focused on automatic identification of metaphorical expressions in text. They used techniques such as supervised classification \cite{mohler-EtAl:2013:Meta4NLP,tsvetkov-mukomel-gershman:2013:Meta4NLP,hovy-EtAl:2013:Meta4NLP}, clustering \cite{ShutovaColing2010,ShutovaNAACL2013}, vector space models \cite{ShutovaCOLING2012,mohler-EtAl:2014:Coling}, lexical resources \cite{krishnakumaran-zhu,wilks-EtAl:2013:Meta4NLP} and web search with lexico-syntactic patterns \cite{VealeHao,LiTACL2013,BollegalaShutova2013}. Two approaches looked explicitly at conceptual metaphor. \newcite{CorMet} automatically acquired domain-specific selectional preferences of verbs, and then, by mapping their common nominal arguments in different domains, arrived at the corresponding metaphorical mappings. 
 \newcite{ShutovaNAACL2013} have previously applied {\hgfc} to acquire a set of metaphorical associations in order to identify metaphorical language in English text. Their intuition was that since metaphorical uses of words constitute a large portion of contexts of abstract nouns in a text corpus, noun clustering techniques are well positioned to identify patterns of metaphorical association.
 In this paper, we apply {\hgfc} to three different languages (English, Russian and Spanish) and investigate its ability to identify cross-corpus and cross-cultural differences in metaphorical framing. We then also investigate the psychological validity of the identified metaphors and differences by conducting their behavioral evaluation.

\section{Experimental data}

\paragraph{English data} The English noun dataset used for clustering contains the 2000 most frequent nouns in the British National Corpus (BNC) \cite{bnc}, which is balanced with respect to topic and genre. 
 The features for clustering were extracted from the English Gigaword corpus \cite{graff2003english} due to its large size. The corpus was parsed using the RASP parser \cite{rasp-2} and \textit{verb--subject, verb--direct object} and \textit{verb--indirect object} relations were then extracted from the parser output. The features used for noun clustering consisted of the verb lemmas occurring in these relations with the nouns in our dataset, indexed by relation type. The feature values were the relative frequencies of the features.


\vspace{-0.1cm}

\paragraph{Spanish data}

The Spanish data was extracted from the Spanish Gigaword corpus \cite{graff2011spanish}. The noun dataset used for clustering consisted of the 2000 most frequent nouns in this corpus. The corpus was parsed using the Spanish Malt parser \cite{Nivre07,Ballesteros2010-SpanishMalt}. \textit{Verb--subject, verb--direct object} and \textit{verb--indirect object} relations were then extracted from the output of the parser and the feature vectors were constructed as in the English system. 

\vspace{-0.1cm}

\paragraph{Russian data}

The Russian data was extracted from the RU-WaC corpus \cite{Sharoff06Wacky}, a two billion-word representative collection of text form the Russian Web. The corpus was parsed using the Russian Malt parser  \cite{SharoffRusMalt}, and \textit{verb--subject, verb--direct object} and \textit{verb--indirect object} relations were extracted to create the feature vectors.  The 2000 most frequent nouns in the RU-WaC constituted the dataset used for clustering.

\section{Method}
\label{sec:HGFC}

 We first cluster nouns using {\hgfc} to create a graph of concepts with different levels of generality. The weights on the edges of the graph indicate the level of association between concepts (represented as clusters). 
{\hgfc} allows us to model multiple relations between concepts simultaneously via soft clustering. This makes it well suited to detect the structure of metaphorical associations, where each concept can be associated with several others.

\subsection{HGFC clustering}

\begin{figure*}[htp]
\begin{center}
\subfigure[]{\label{fig:bipart-a}\includegraphics[scale=0.5]{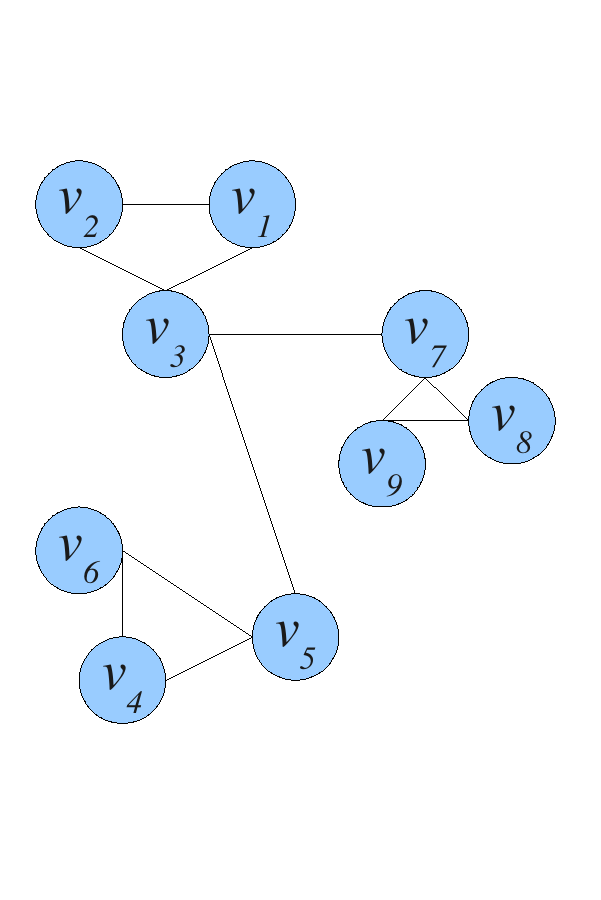}}
\subfigure[]{\label{fig:bipart-b}\includegraphics[scale=0.5]{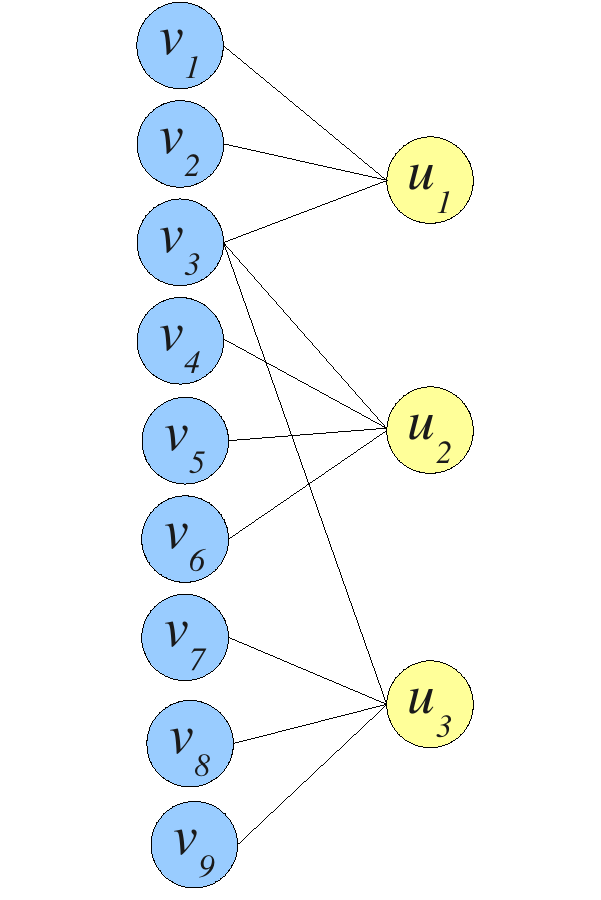}}
\subfigure[]{\label{fig:bipart-c}\includegraphics[scale=0.5]{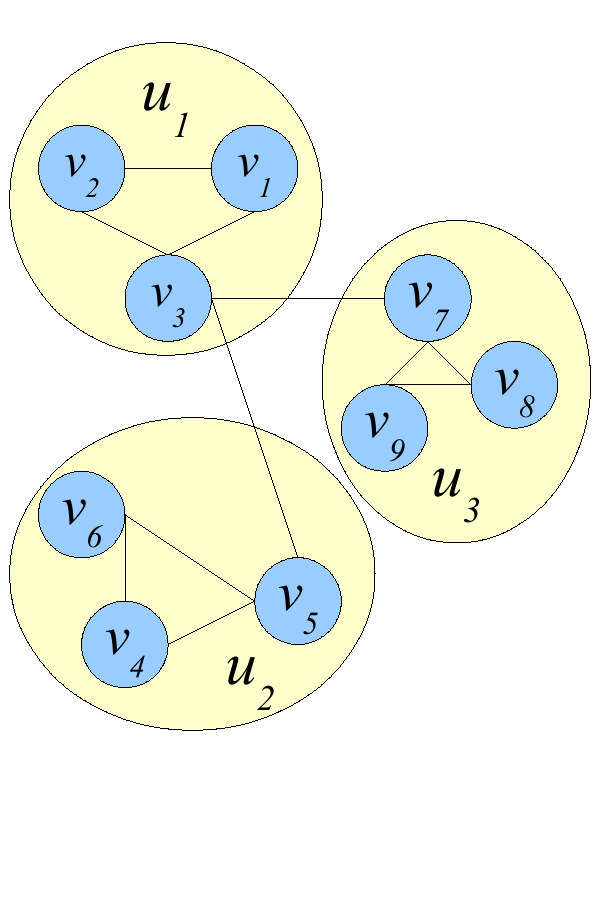}}
\subfigure[]{\label{fig:bipart-d}\includegraphics[scale=0.5]{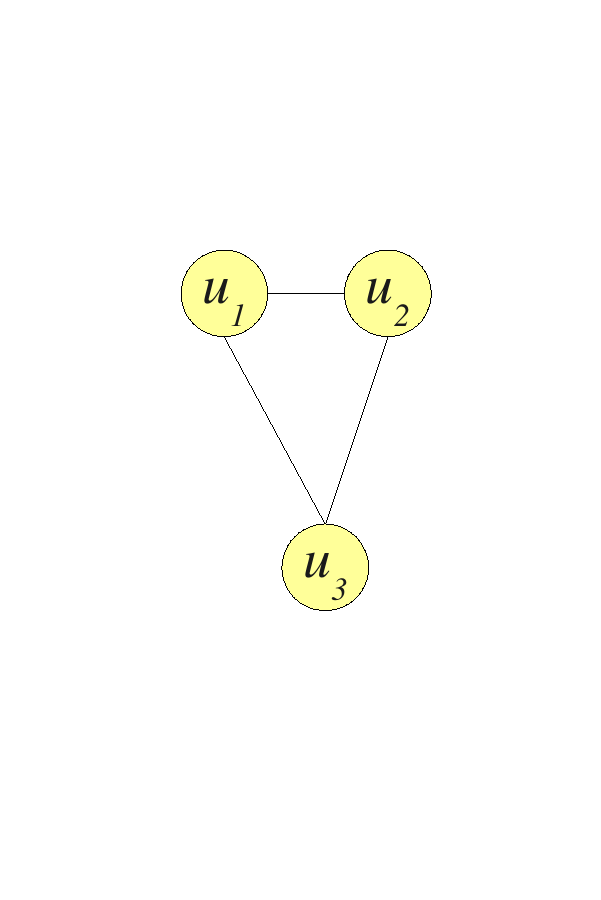}}
\subfigure[]{\label{fig:bipart-e}\includegraphics[scale=0.5]{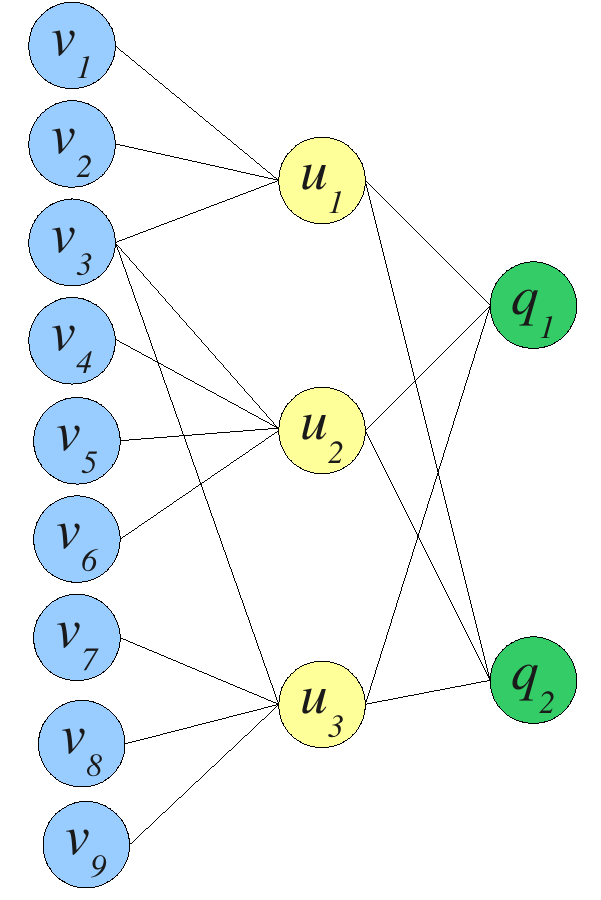}}
\end{center}
\label{fig:bipart}
\vspace{-0.2in}
\caption{(a) An undirected graph $G$ representing the similarity matrix; (b) The bipartite graph showing three clusters on $G$; (c) The induced clusters $U$; (d) The new graph $G_1$ over clusters $U$; (e) The new bipartite graph over $G_1$ }
\vspace{-0.4cm}
\end{figure*}

The algorithm successively derives probabilistic bipartite graphs for every level in the hierarchy. 
 Given a set of nouns, $V=\{v_n\}_{n=1}^{N}$, we first construct their similarity matrix $W$ using Jensen-Shannon Divergence as a measure. The matrix $W$ encodes an undirected similarity graph $G$, where the nouns are mapped to vertices and their similarities represent the weights $w_{ij}$ on the edges between vertices $i$ and $j$ (see Fig. \ref{fig:bipart-a}). The clustering problem can now be formulated as partitioning of $G$. 

The graph $G$ and the cluster structure can be represented by a bipartite graph $K(V,U)$, where $V$ are the vertices on $G$ and $U=\{u_p\}_{p=1}^{m}$ represent $m$ hidden clusters. For example, 
$V$ on $G$ can be grouped into three clusters $u_1$, $u_2$ and $u_3$ (Fig. \ref{fig:bipart-b}). The matrix $B$ denotes the $n \times m$ adjacency matrix, with $b_{ip}$ being the connection weight between the vertex $v_i$ and the cluster $u_p$.  Thus, $B$ represents the connections between clusters at an upper and lower level of clustering. In order to derive the clustering structure, we first need to compute $B$ from the original similarity matrix. The similarities $w_{ij}$ in $W$ can be interpreted as the probabilities of direct transition between $v_i$ and $v_j$: $w_{ij}=p(v_i,v_j)$. The bipartite graph $K$ also induces a similarity ($W'$) between $v_i$ and $v_j$, with all the paths from $v_i$ to $v_j$ going through vertices in $U$. This means that the similarities $w'_{ij}$ are to be computed via the weights $b_{ip} = p(v_i, u_p)$. 

\vspace{-0.1in}
\small
\begin{equation}
\begin{gathered}
p(v_i, v_j) = p(v_i) p(v_j | v_i) = p(v_i) \sum_p p(u_p|v_i) p(v_j|u_p) \\ =p(v_i) \sum_p \frac {p(v_i, u_p) p(u_p, v_j)} {p(v_i) p (u_p)} = \sum_p \frac{b_{ip} b_{jp}}{\lambda_p},
\end{gathered}
\end{equation}
\normalsize
where $\lambda_i = \sum_{i=1}^{n}b_{ip}$ is the degree of vertex $u_p$. The similarity matrix $W'$ can thus be derived as:

\vspace{-0.1in}
\small
\begin{equation}
\label{eq:Wprime}
 W': w_{ij}'  =\sum_{p=1}^{m}\frac{b_{ip}b_{jp}}{\lambda_p} = (B \Lambda^{-1} B^T)_{ij}, 
\end{equation}
\normalsize
where $\Lambda  = \text{diag}(\lambda_1,...,\lambda_m)$.  $B$ can then be found by minimizing the divergence distance ($\zeta$) between the similarity matrices $W$ and $W'$.

\vspace{-0.1in}
\small
\begin{equation}
\label{eq:objective}
 \min_{H,\Lambda} \zeta(W,H\Lambda H^T) , s.t. \sum_{i=1}^{n}h_{ip}=1 
\end{equation}
\normalsize
We remove the coupling between $B$ and $\Lambda$ by setting $H=B\Lambda^{-1}$. Following \newcite{yu2006soft} we define $\zeta(X,Y) = \sum_{ij}(x_{ij}\log\frac{x_{ij}}{y_{ij}}-x_{ij}+y_{ij})$. 
 \newcite{yu2006soft} showed that this cost function is non-increasing under the following update rule: 
 
\vspace{-0.1in}
\small
\begin{equation}
\label{eq:h}
 \tilde{h}_{ip} \propto h_{ip} \sum_j\frac{w_{ij}}{(H\Lambda H^T)_{ij}}\lambda_p h_{jp} 
\text{ s.t.}  \sum_{i}\tilde{h}_{ip}=1
\end{equation}
\begin{equation}
\label{eq:lambda}
\tilde{\lambda}_{p} \propto \lambda_{p} \sum_j\frac{w_{ij}}{(H\Lambda H^T)_{ij}}h_{ip} h_{jp}
\text{ s.t.} \sum_{p}\tilde{\lambda}_{p}=\sum_{ij}w_{ij} 
\end{equation}
\normalsize
We optimized $\zeta$ by alternately updating $h$ and $\lambda$. 

A flat clustering algorithm can be induced by computing $B$ and assigning a lower level node to the parent node that has the largest connection weight. The number of clusters at any level can be determined by counting the number of non-empty nodes. To create a hierarchical graph we need to repeat the above process to successively add levels of clusters to the graph. To create a bipartite graph for the next level, we first need to compute a new similarity matrix for the clusters $U$. Similarity between clusters $p(u_p,u_q)$ can be induced from $B$:

\vspace{-0.15in}
\small
\begin{equation}
\label{eq:pu} 
p(u_p,u_q)  =  p(u_p)p(u_p|u_q) = (B^T D^{-1} B)_{pq}  
\end{equation}
\normalsize
where $D = \text{diag}(d_1,...,d_n); d_i = \sum_{p=1}^{m}b_{ip}$.
We can then construct a new graph $G_1$ (Fig. \ref{fig:bipart-d}) with the clusters $U$ as vertices, and the cluster similarities $p(u_p,u_q)$ as the connection weights. 
The clustering algorithm can now be applied again (Fig. \ref{fig:bipart-e}). This process can go on iteratively, leading to a hierarchical graph. 

The number of levels ($L$) and the number of clusters ($m_\ell$) are detected automatically, using the method of \newcite{SunKorhonen2011}. Clustering starts with an initial setting of the number of clusters $m_1$ for the first level. In our experiments, we set the value of $m_1$ to 800. For the subsequent levels, $m_\ell$ is set to the number of non-empty clusters on the parent level -- 1.  The matrix $B$ is initialized randomly and its rows are then normalized. 

For a word $v_i$, the probability of assigning it to cluster $x_{p}^{(\ell)} \in X_\ell $ at level $\ell$ is given by:

\vspace{-0.15in}
{\small
\begin{eqnarray}
p(x_p^{(\ell)}|v_i) &  = & \sum_{X_{\ell-1}} ... \sum_{x^{(1)} \in X_{1}} p(x_p^{(\ell)}|x^{(\ell-1)})... p(x^{(1)}|v_i) \nonumber \\
& = & (D_1^{(-1)}B_1D_2^{-1}B_2 ... D_\ell^{-1}B_\ell)_{ip}
\label{eq:p}
\end{eqnarray}
}
\vspace{-0.2in}

\noindent  \newcite{SunKorhonen2011} have shown that  $m_\ell$ is non-increasing for higher levels. The algorithm can thus terminate when all nouns are assigned to one cluster. We run 1000 iterations of updates of $h$ and $\lambda$ (eq. \ref{eq:h} and \ref{eq:lambda}) for each two adjacent levels. The algorithm can be summarized as follows:
\begin{algorithm}
\begin{algorithmic}
\small
\State\hspace{\algorithmicindent}\textbf{Require:} {$N$ nouns $V$, initial number of clusters $m_1$.}

      \State\hspace{\algorithmicindent}  Compute the similarity matrix $W_0$ from $V$.
      \State\hspace{\algorithmicindent}  Build the graph $G_0$ from $W_0$, $\ell \gets 1$.
      \State\hspace{\algorithmicindent}\textbf{while}{ $m_\ell>1$ }\textbf{do}
           \State\hspace{\algorithmicindent}\hspace{\algorithmicindent} Factorize $G_{\ell-1}$ to obtain bipartite graph $K_{\ell}$ with adjacency matrix $B_\ell$ (eqs. \ref{eq:h}, \ref{eq:lambda}).
           \State\hspace{\algorithmicindent}\hspace{\algorithmicindent} Build a graph $G_\ell$ with similarity matrix $W_\ell=B_\ell^T D_\ell^{-1} B_\ell$ according to equation \ref{eq:pu}.
           \State\hspace{\algorithmicindent}\hspace{\algorithmicindent} $\ell \gets \ell+1$  ; $m_\ell \gets$ No. non-empty clusters --1.
     \State\hspace{\algorithmicindent}\textbf{end while} 
\State\hspace{\algorithmicindent}\Return $B_\ell, B_{\ell-1} ... B_1$
 \end{algorithmic}
\end{algorithm}


The resulting graph is composed of a set of bipartite graphs defined by $B_\ell, B_{\ell-1}, ..., B_1$. 
For a given noun, we can rank the clusters at any level according to the soft assignment probability (eq. \ref{eq:p}). The clusters that have no member noun were hidden from the ranking since they do not explicitly represent any concept. However, these clusters are still part of the organisation of the conceptual space and contribute to the probability for the clusters at upper levels (eq. \ref{eq:p}).  We call the view of the hierarchical graph where these empty clusters are hidden an {\it explicit graph}.

\subsection{Identifying metaphorical associations}
\label{sec:MetAss}

Once we obtained the explicit graph of concepts, we can identify metaphorical associations based on the weights on the edges of the graph. 
 Taking a single noun (e.g. \textit{fire}) as input, the system computes the probability of its cluster membership for each cluster at each level, using these weights (eq. \ref{eq:p}). We expect the cluster membership probabilities to indicate the level of association of the input noun with the clusters. The system then ranks the clusters at each level based on these probabilities. We chose level 3 as the optimal level of generality based on our qualitative analysis of the graph. The system selects 6 top-ranked clusters from this level 
  and excludes the literal cluster containing the input concept (e.g. ``\textit{fire flame blaze}''). The remaining clusters represent target concepts associated with the input concept.
 
\begin{figure}[t]
\centering
\fbox{\begin{minipage}{3.05in}
\small{
\textbf{\textsc{source}: fire}

   \textsc{target 1}: sense hatred emotion passion enthusiasm hope feeling  optimism hostility excitement anger ...
   
   \textsc{target 2}:  violence fight resistance clash rebellion battle  fighting riot revolt war confrontation revolution ...
   
   \textsc{target 3}: alien immigrant
   
   \textsc{target 4}: prisoner hostage inmate
\vspace{-0.2cm}


\rule{3.05in}{.1pt}

\textbf{\textsc{source}: disease}

   \textsc{target 1}: fraud outbreak offense crime violation abuse conspiracy corruption terrorism suicide ... 
   
   \textsc{target 2}: opponent critic rival
   
   \textsc{target 3}: execution destruction signing
   
   \textsc{target 4}: refusal absence fact failure lack delay
   
}
\end{minipage}}
\caption{Metaphors identified in the English data}
\label{fig:MetAssoc}
\vspace{-0.5cm}
\end{figure}

\begin{figure}[t]
\centering
%
%
%
%
%
%
%
%
%
%
%
%
%
%

\includegraphics[width=3.2in]{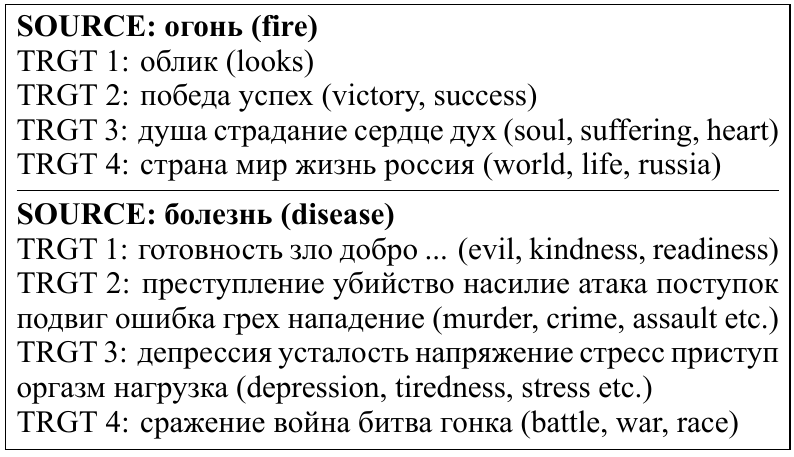}
\caption{Metaphors identified in the Russian data}
\label{fig:MetAssocRU}
\vspace{-0.4cm}
\end{figure}

  Example output for the input concepts of \textit{fire} and \textit{disease} in English is shown in Fig.~\ref{fig:MetAssoc}. One can see that each noun-to-cluster mapping represents a new conceptual metaphor, e.g. \textsc{emotion} is \textsc{fire}, \textsc{violence} is \textsc{fire}, \textsc{crime} is a \textsc{disease}. These mappings are exemplified in language by numerous metaphorical expressions (e.g. ``his anger \textit{blazed}'', ``violence \textit{flared}''). Figs. \ref{fig:MetAssocES} and \ref{fig:MetAssocRU} show metaphorical associations identified in the Spanish and Russian data for the same source concepts. One can see that \textsc{feelings} are associated with \textsc{fire} in all three languages. However, many of the identified metaphors differ across languages: e.g., \textsc{victory, success} and \textsc{looks} are viewed as \textsc{fire} in Russian, while \textsc{immigrants} and \textsc{prisoners} are associated with \textsc{fire} in English and Spanish.
All of the languages exhibit \textsc{crime is a disease} metaphor, with Russian and Spanish generalising it to \textsc{violence is a disease}. 
 While we do not claim that this output is exhaustively representative of all conceptual metaphors present in a particular culture, we believe that these examples showcase some interesting differences in the use of metaphor across datasets that can be discovered by our method.

\begin{figure}[t]
\centering
\fbox{\begin{minipage}{3.1in}
\small{
\textbf{\textsc{source}: fuego (fire)}

   \textsc{trgt 1}: esfuerzo negocio tarea debate operaci\'{o}n operativo ofensiva gira acci\'{o}n actividad  campa\~{n}a gesti\'{o}n ... 
    
   \textsc{trgt 2}: quiebra indignaci\'{o}n ira  p\'{a}nico caos alarma ... 
   
   \textsc{trgt 3}: reh\'{e}n refugiado preso prisionero inmigrante ... 
   
   \textsc{trgt 4}: soberan\'{i}a derecho independencia libertad ...
\vspace{-0.2cm}

\rule{3.05in}{.1pt}

\textbf{\textsc{source}: enfermedad (disease)}

   \textsc{target 1}: calentamiento inmigraci\'{o}n impunidad
      
   \textsc{target 2}:  desaceleraci\'{o}n brote fen\'{o}meno epidemia sequ\'{i}a violencia mal recesi\'{o}n escasez contaminaci\'{o}n
      
    \textsc{target 3}:  petrolero fabricante gigante firma aerol\'{i}nea
      
    \textsc{target 4}:  mafia

}
\end{minipage}}
\caption{Metaphors identified in the Spanish data}
\label{fig:MetAssocES}
\vspace{-0.3cm}
\end{figure}

\section{Evaluation within languages}

We first evaluated the quality of metaphor identification in individual languages. As there is no comprehensive gold standard of metaphorical mappings available, we evaluated the identified mappings against human judgements. 

\vspace{+0.2cm}
\noindent \textbf{Baseline}  \hspace{+0.1cm} We compared the system performance to that of an agglomerative clustering baseline (\textsc{agg}). We constructed \textsc{agg} using SciPy implementation  \cite{scipy} of Ward's linkage method \cite{ward1963hierarchical}. The output tree was cut according to the number of levels and clusters in the explicit {\hgfc} graph. We converted this tree into a graph by adding connections from each cluster to all the clusters one level above. We computed the connection weights as cluster distances measured using Jensen-Shannon Divergence between the cluster centroids. 
 This graph was then used in place of the {\hgfc} graph. 

\vspace{+0.2cm}
\noindent \textbf{Evaluation setup and results}  \hspace{+0.1cm} To create our dataset, we extracted 10 common source concepts that map to multiple targets from the Master Metaphor List \cite{MastMetList} and linguistic analyses of metaphor \cite{ShutovaLREC}. These included \textsc{fire, child, speed, war, disease, breakdown, construction, vehicle, system, business}. We then translated them into Spanish and Russian. Each of the systems identified 50 mappings for the given source domains. This resulted in a set of 100 conceptual metaphors for each language. Each of them represents a number of submappings since the target concepts are clusters of nouns. These were then evaluated against human judgements in two different experimental settings. 

\vspace{+0.1cm}
\noindent \underline{Setting 1 (precision-oriented)}:

\noindent  The judges were presented with a set of mappings identified by the system and the baseline, randomized.  They were asked to annotate the mappings they considered valid as correct. A mapping was to be considered valid if it could be exemplified by a metaphorical expression. 
  
  Two judges per language, who were native speakers of English, Russian and Spanish participated in this experiment. All of them held at least a Bachelor degree.   Their agreement was measured at $\kappa =0.60$ for English, $\kappa =0.59$ for Spanish, and $\kappa =0.55$ for Russian. The main differences in the annotators' judgements stem from the fact that some metaphorical associations are less obvious and common than others, and thus need more context (or imaginative effort) to establish. Such examples of disagreement included the metaphorical mappings \textsc{intensity} is \textsc{speed},  \textsc{goal} is a \textsc{child}, \textsc{collection} is a \textsc{system}, \textsc{illness} is a \textsc{breakdown}. 

The system performance was then evaluated against these judgements in terms of precision ($P$), i.e. the proportion of the valid metaphorical mappings among those identified. We calculated system precision as an average over both annotations in a given language. The results are presented in Table~\ref{tab:MetAssocResults}.

\begin{table}[t]
\centering
    \begin{tabular}{lllll}
\hline
&\textsc{agg} $P$&\textsc{agg} $R$&\textsc{hgfc} $P$&\textsc{hgfc} $R$\\
\hline
EN  & 0.36 & 0.11&0.69 & 0.61\\
ES  & 0.23 &0.12& 0.59&0.54\\
RU  & 0.28 &0.09& 0.62&0.42\\
  \end{tabular}
  \caption{\textsc{hgfc} and baseline performance}
  \label{tab:MetAssocResults}
  \vspace{-0.4cm}
\end{table}

\vspace{+0.1cm}
\noindent \underline{Setting 2 (recall-oriented)}: To measure recall, $R$, of the systems we asked two annotators per language (native speakers with a background in metaphor, different from Set. 1) to write down up to 5 target concepts they strongly associated with each of the 10 source concepts. Their annotations were then aggregated into a single metaphor association gold standard. The gold standard consisted of 63 mappings for English, 70 mappings for Spanish and 68 mappings for Russian. The recall of the systems, as measured against this gold standard, is shown in Table~\ref{tab:MetAssocResults}. 

\vspace{+0.2cm}
\noindent \textbf{Discussion and error analysis}  \hspace{+0.1cm} \textsc{hgfc} outperforms the \textsc{agg} baseline in all evaluation settings and identifies valid metaphorical associations for a range of source concepts.  \textsc{agg}, although less suitable for the task, still identified a number of interesting mappings missed by {\hgfc} (e.g. \textsc{career} is a \textsc{child, language} is a \textsc{system, corruption} is a \textsc{vehicle}) and a number of mappings in common with {\hgfc} (e.g. \textsc{debate} is a \textsc{war, destruction} is a \textsc{disease}). The fact that both \textsc{hgfc} and \textsc{agg} identified valid metaphorical mappings across languages confirms our hypothesis that clustering techniques are well suited to detect metaphorical patterns in a distributional word space in principle. 


 The most frequent type of error of \textsc{hgfc} across the three languages is the presence of target clusters similar or closely related to the source noun. For instance, the source noun \textsc{child} tends to be linked to other "human" clusters across languages, e.g. the \textit{parent} cluster in English, the \textit{student, resident} and \textit{worker} clusters in Spanish and the \textit{crowd, journalist} and \textit{emperor} clusters in Russian. 

The performance of the Russian and the Spanish systems is slightly lower than that of the English system. This may be due to errors from the data preprocessing step, i.e. parsing. Parsing quality in English is likely to be higher than in Russian or Spanish, for which fewer parsers exist. Another important difference lies in the corpora used. While the English and Spanish systems were applied to the Gigaword corpora (containing data from news sources), the Russian system was applied to the Web data containing noisier text (including misspellings, slang etc.)

\section{Cross-linguistic analysis}

We then investigated the differences in metaphorical framing, as identified by our systems across languages. We ran the systems with a larger set of source domains taken from the literature on metaphor and conducted a qualitative analysis of the resulting metaphorical mappings. As one might expect, the majority of the identified mappings are present across languages. For instance, 
\textsc{debate} or \textsc{argument} are associated with \textsc{war} in all three languages;
\textsc{crime} is universally associated with \textsc{disease} and \textsc{money} with \textsc{liquid} etc. 

Importantly, our methods were also able to capture differences in metaphorical framing in the three languages. For instance, they exposed some interesting differences in the domains of \textit{business} and \textit{finance}.
The Spanish data manifested rather negative metaphors about business, market and commerce: \textsc{business} was typically associated with \textsc{bomb, fire, war, disease} and \textsc{enemy}. While it is the case that \textsc{business} is typically discussed in terms of a \textsc{war} or a \textsc{race} in English and Russian, the other four Spanish metaphors are uncommon. Russian, in fact, has rather positive metaphors for the related concepts of \textsc{money} and \textsc{wealth}, which are strongly associated with \textsc{sun, light, star} and \textsc{food}, possibly indicating that money is viewed primarily as a way to improve one's life. In contrast, in English, \textsc{money} is frequently discussed as a \textsc{weapon} -- a means to achieve a goal or win a struggle (related to \textsc{business is a war} metaphor). At the same time, the English data exhibits positive metaphors for \textsc{power} and \textsc{influence}, which are viewed as \textsc{light, sun} or \textsc{wing}. In Russian, on the contrary, \textsc{power} is associated with \textsc{bomb} and \textsc{bullet}, perhaps linking it to the concepts of physical strength and domination. Yet, the concepts of \textsc{freedom} and \textsc{independence} were also associated with a \textsc{wing, weapon} and \textsc{strength} in the Russian data. 
  English data exhibited more negative metaphors for \textit{immigration} than Russian or Spanish, with \textsc{immigrants} viewed as \textsc{fire} or \textsc{enemies}, possibly indicating danger. 
  
  While the above differences may be a direct result of the contemporary socio-economic context and political rhetoric, and are likely to change over time, other conceptual differences have a deeper grounding in our culture and the way of life. For instance, the concept of \textsc{birth} tends to be strongly associated with \textsc{light} in Spanish and \textsc{battle} in Russian, each metaphor highlighting a different aspect of birth. Another interesting difference concerned the framing of the concept of \textit{economy} in English and Spanish. In English data, \textsc{economy} is viewed predominantly as a \textsc{vehicle}\footnote{Although other metaphors for the \textit{economy} exist in English, the system identifies the statistically dominant ones.}, that can be \textit{driven forward} or \textit{slowed down}. In Spanish, on the contrary, the economy is thought of in terms of its \textsc{size} and \textsc{growth}, but not motion. 
Research in cognitive psychology  \cite{CasasantoBoroditsky2008,Fuhrman2011} suggests that such cross-linguistic differences in conventionalised metaphors have significance beyond language and can be associated with contrastive behavioural patterns across the different linguistic communities. In the next section, we present a behavioural study aimed at assessing the psychological validity of a subset of cross-linguistic differences identified by our model.

\section{Behavioural evaluation}

We focused on the difference in the metaphors used by English vs. Spanish speakers when discussing changes in the economy.  The observed difference may be a property of language or it could also reflect entrenched conceptual differences. In order to investigate this, we test whether patterns of behavior consistent with this difference in metaphorical framing arise cross-linguistically in response to non-linguistic stimuli. 

\subsection{Experimental setup}

We recruited 60 participants from one English-speaking country (USA) and 60 participants from three Spanish-speaking countries (Chile, Mexico, Spain) using the CrowdFlower  platform. Participants first read a brief description of the task, which introduced them to a fictional country in which economists are devising a graphic for representing changes in the economy. 
 They then completed a demographic questionnaire including information about their native language. Results from 9 US and 3 non-US participants were discarded for failure to meet the language requirement.

Participants navigated to a new page to complete the experimental task. Stimuli were presented in a  $1200 \times 700$-pixel frame. The center of the frame contained a sphere with a 64-pixel diameter. For each trial, participants clicked on a button to activate an animation of the sphere which involved (1) a positive displacement (in rightward pixels) of 10\% or 20\%, or a negative displacement (in leftward pixels) of 10\% or 20\%; and, (2) an expansion (in increased pixel diameter) of 10\% or 20\%, or a contraction (in decreased pixel diameter) of 10\% or 20\%.\footnote{The English experimental interface can be viewed at \url{http://goo.gl/W3YVfC}.
The Spanish interface is identical, but for a direct translation of the guidelines.} They were then asked to judge whether the economy has "improved" or "worsened" based on the graphic.


Participants saw each of the resulting conditions 3 times. The displacement and size conditions were drawn from a random permutation of 16 conditions using a Fisher-Yates shuffle \cite{FisherYates}. Crucially, half of the stimuli contained conflicts of information with respect to the size and displacement metaphors for economic change (e.g. the sphere could both grow and move to the left). 
Overall we expected the Spanish speakers' responses to be more closely associated with changes in diameter (due to the salience of the \textit{size} metaphor) and the English speakers' responses with displacement (due to the salience of the \textit{vehicle} metaphor). We expected these differences to be most prominent in the conflicting trials, which force the participants to choose between the two metaphors. We focus on these conflicting trials in our analysis.

\subsection{Results}

In trials where stimuli moving rightward were simultaneously contracting, English and Spanish speakers responded that the economy improved 66\% and 43\% of the time respectively. 
 In trials where stimuli moving leftward were simultaneously expanding, English and Spanish speakers judged the economy to have improved 34\% and 55\% of the time respectively. 
 These results indicate that English speakers judgments were more biased towards changes in the sphere's displacement, while Spanish speakers judgments towards changes in diameter. These results support our expectation on the relevance of different metaphors when reasoning about the economy by the English and Spanish speakers.
 

To examine the significance of these effects, we fit a binary logit mixed effects model \cite{FoxWeisberg} to the data. The full analysis modeled judgment with native language, displacement, and size as fully crossed fixed effects and participant as a random effect. This analysis confirmed that native language was associated with participants' judgments about economic change. It indicated that changes in size affected English and Spanish speakers' judgments differently ($p<0.001$), with an increase in size increasing the odds ($e^\beta=2.5$) of a judgment of \textit{Improved} by Spanish speakers and decreasing the odds ($e^\beta=0.44$) of a judgment of \textit{Improved} by English speakers. A Type II Wald test revealed the interaction between language and size to be highly statistically significant ($\chi^2(1)<0.001$).

\section{Conclusion}

We presented a method that identifies patterns of metaphorical framing in a large text corpus. Despite being fully unsupervised, it operates with an ecouraging precision. It is portable across datasets and languages, and discovers interesting cross-linguistic differences in metaphorical framing. We have shown that the method predicts patterns consistent with behavioural data. While much territory remains to be investigated with respect to delimiting the nature of this relationship, these results represent a first step toward establishing an association between information mined from large textual data collections and information observed through behavioural responses on a human scale.

\bibliographystyle{emnlp2016}

\bibliography{CL-paper}

\end{document}